\documentclass{IOS-Book-Article}

\usepackage{hyperref}
\usepackage{graphicx}
\usepackage{mathptmx}
\usepackage{soul}\setuldepth{article}
 
\def\hb{\hbox to 10.7 cm{}}

\usepackage{color}

\begin{document}

\pagestyle{headings}
\def\thepage{}

\begin{frontmatter}              

\title{Classifying concepts via visual properties}


\author{\fnms{Fausto} \snm{Giunchiglia}}
and
\author{\fnms{Mayukh} \snm{Bagchi}}

\address{Department of Information Engineering and Computer Science (DISI), \\ University of Trento, Italy.}

\begin{abstract}
We assume that \textit{substances} in the world are represented by two types of concepts, namely \textit{substance concepts} and \textit{classification concepts}, the former instrumental to (visual) perception,  the latter to (language based) classification. 
Based on this distinction, we introduce a general methodology for building lexico-semantic hierarchies of substance concepts, where nodes are  annotated with the media, e.g., videos or photos, from which substance concepts are extracted, and are associated with the corresponding classification concepts. The methodology is based on Ranganathan's original faceted approach, contextualized to the problem of classifying substance concepts. The key novelty is that the hierarchy is built exploiting the \textit{visual} properties of substance concepts, while the \textit{linguistically defined} properties of classification concepts are only used to describe substance concepts.
The validity of the approach is exemplified by providing some highlights of an ongoing project whose goal is to build a large scale multimedia multilingual concept hierarchy.
\end{abstract}

\end{frontmatter}

\section{Introduction}
\label{sec:Introduction}

Concepts are a foundational notion for any theory of mind, no matter whether these theories are more theoretically oriented (as in, e.g., the Philosophy of Mind, Lexical Semantics), or more application oriented, (as in, e.g., Information Systems,  Artificial Intelligence (AI), Computational Linguistics). As of to day, it is somehow universally accepted that concepts are mental representations of substances, as they occur in the world, but what these representations are, what they are for, and how many of them exist, these are all questions for which there is no uniform view.
So far, the mainstream approach, mainly but not only, in the application oriented work, has been to model concepts as \textit{classes}, which in turn are populated by \textit{instances}.  This approach was termed \textit{Descriptionism} in \cite{millikan2000} to emphasize the fact that, in this work, the main goal is \textit{to describe} what is the case in the world, this description being instrumental to the intended (specific) use of concepts, i.e., that of providing an account of and also implementing  phenomena such as knowledge acquisition and representation, reasoning, natural language processing and classification.

In earlier work \cite{giunchiglia2016}, the authors propose a complementary view, building upon some recent results in the field of Teleosemantics \cite{macdonald2006}, previously called Biosemantics \cite{millikan1989}, and in particular of the work of Ruth Millikan \cite{millikan2000,millikan2004,millikan2005}. In this work, concepts are seen as \textit{(etiological) functions}. Here, the notion of etiological function must be read as meaning  \emph{`intended for'} or \emph{`devised for'} with respect to a referent device \cite{neander1991}, this notion being adapted from the notion of biological causation. To exemplify, the occipital lobe of our brain is devised for visual perception, where the occipital lobe is the referent device `intended to' perform the function of vision. 
As from \cite{giunchiglia2016}, this view allows us to distinguish between two types of concepts, as follows:

\begin{enumerate}
    \item \textit{substance concepts}, namely mental representations which support the \textit{recognition of what is the case in the world},\footnote{We restrict therefore ourselves to substances which can be perceived, e.g., \textit{objects}.} and 
    \item \textit{classification concepts}, namely mental representations which support the \textit{classification of what has been recognized as substance concepts}, classification being the key activity which allows to talk and reason about substance concepts.
\end{enumerate}
   Classification concepts capture the Descriptionistic view of concepts as classes. The more novel notion  of substance concepts has been successfully tested in the development of a system capable of performing human-like object recognition \cite{ceor,giunchiglia2021a}. Notice that we assume the existence of two types of  concepts, where substance concepts and classification concepts perform different functions and are built following independent processes.\footnote{It is worth noticing how, according to the work in Neuroscience, the brain actually holds multiple representations of the world (at least one per sense) \cite{martin2001,NSenyclo}.} The problem which arises is how to make sure that, given a certain substance in the real world, its substance concept and classification concept mental representations both encode the substance itself. The work in \cite{ceor,giunchiglia2021a} provides a general solution to this problem by ensuring that: (i) there is one-to-one correspondence between the two types of concepts; (ii) substance concepts are organized into hierarchy based on \textit{Genus-Differentia} intensional definitions, as it is usually the case in Lexical Semantics \cite{parry1991,miller1990}, and (iii) the one-to-one mapping is actually correct in the sense that the selected classification concept properly names and describes the substance been perceived. 

 The goal in this paper is to provide a general methodology for building such an integrated hierarchy of classification and substance concepts. The ultimate goal is to use the proposed methodology to develop a lexico-semantic hierarchy where classification concepts are annotated with media, e.g., photos or videos, representing the corresponding substance concepts. The approach we propose is based on Ranganathan's faceted methodology for building conceptual hierarchies \cite{srr67}. 
 %
%
The main novelty is that, contrarily to what has been the case so far, see, e.g., all the work in the construction of lexico-semantic hierarchies \cite{miller1990,fausto2017}, the hierarchies are built based on the visual properties of substance concepts, rather than by exploiting the linguistically defined properties of classification concepts. The properties of classification concepts are \textit{only} used to describe substances, as perceptually represented by substance concepts and, accordingly, organized in hierarchies. This fact, in turn enables the definition of a very compelling set of rules which allow for the construction of high quality classification hierarchies which at the same time encode the visual properties of substances, as well as their linguistically defined properties, thus being amenable for multi-media classification.

The paper is structured as follows. In Section \ref{sec:A Teleological View of Concepts} we provide a short synthesis of the teleological view of substance and classification concepts. In Section \ref{sec:A Knowledge Organization View of concepts} we introduce Ranganathan's  faceted view of concepts. Here we also show how Ranganathan's approach naturally maps to the proposed etiological view of concepts. In Section \ref{sec:A Canonical Framework for Concept Hierarchies} we introduce a set of canons, adapted from Ranganathan's faceted approach, which prescribe how to build high quality hierarchies. In Section \ref{sec:A Multimedia UKC} we provide some insights about how the ideas presented in this paper can be exploited in the construction of a multi-media lexico-semantic hierarchy. Finally, Section \ref{sec:Conclusion} concludes the paper.

\section{A Teleological View of Concepts}
\label{sec:A Teleological View of Concepts}

We assume that the world is populated by \emph{substances} \cite{millikan2000,giunchiglia2016}, which are subjects of knowledge amenable to perception, where, in the parlance of Teleosemantics, substances are \emph{``those things about which you can learn from one encounter something of what to expect on other encounters, where this is no accident but the result of a real connection”} \cite{millikan2000}. Substances are of two types: (i) \emph{individuals}, which are \emph{representations of sets of temporal occurrences of substances} commonly identified by employing proper nouns (such as the Colosseum, my pet fish which I named Oreo), and (ii) \emph{real kinds}, which \emph{``allow successful inductions to be made from one or a few members to other members of the kind not by accident”} \cite{millikan2000} (such as gold, fish, professor). The recognition of such substances happens via \emph{sets of encounters}, with encounters being events through which substances partially exhibit themselves  to perception over \emph{space-time}. In this context, it is important to note that the uniquely identifying characteristic(s) which facilitate the recognition of a substance, as distinct from other substances, is grounded in its \emph{substance causal factor}. Here the causal factor is meant to be a set of inner characteristics which are distinctive of the substance and are directly responsible for its broad \emph{invariance} across encounters (for instance, homeostasis is a causal factor in living beings). Such invariance, in due course, is manifested as a set of \emph{outer characteristics} which are discriminating stimuli which allow to uniquely perceive the substance (e.g., color, shape). 

The incremental generation of knowledge from such acts of perception is enabled by \emph{substance concepts}, which we characterize via their \emph{recognition abilities}, viz. \emph{``abilities which allow ... to realize that the substance involved in the current encounter is the same substance as from previous encounters"} \cite{giunchiglia2016}. Substance concepts enable functions which facilitate the visual recognition of substances within the same encounter and across encounters. Thus, substance concepts are defined by a set of visual characteristics (i.e, substance properties), called \textit{visual objects}, where visual objects are sequences of similar \textit{visual frames} \cite{ceor,giunchiglia2021a}. Via visual frames it is thus possible to compute the visual (dis)similarity among what is perceived in different encounters. A concrete example would be, for instance, the incremental generation of knowledge about the substance concept ``fish"\footnote{Throughout the paper, to distinguish between substance concepts and substances, we write the former in ``quotes". For example, fish is a substance whereas ``fish" is a substance concept.} as recognized from a set of encounters with the substance fish (oblivious of image background and composition). In such sequences the fish could be described, for instance, by visual objects which depict how it appears from the front, from the side, at night, in dark water and so on.

The fact that classification, the ability of \emph{``reducing the load on memory, and of helping us to store and retrieve information efficiently"} \cite{giunchiglia2016}, is distinct from recognition is reinforced by the very nature of representations of the (part of the) world they generate - the former rooted in lexical semantics \cite{miller1990} and the later grounded in perception \cite{millikan2000}. \emph{Classification Concepts} encode concepts, in Lexical Semantics jargon, \textit{word senses}, and thereby model the diversity of the world in terms of \emph{classes} (i.e., sets of instances lexicalized as nouns) and corresponding \emph{properties}. These are, thus, abilities functioning towards \emph{``organizing instances into classes as a function of their properties"} \cite{giunchiglia2016} in the form of lexical-semantic (classification) hierarchies. Such hierarchies are built following the intensional paradigm of \emph{Genus-Differentia} \cite{parry1991}. Here \emph{Genus} refers to an existing intensional definition characterized by a shared set of properties across distinct objects, for instance, the common linguistically defined properties of a trout fish, e.g., its color, size, weight, movements, while \emph{Differentia} implies a set of novel properties, different from the ones of \emph{Genus}, utilized for discriminating amongst objects with the same genus. One such example would be the set of  distinguishing properties for \emph{rainbow trout}, \emph{steelhead trout} and \emph{Dolly Varden trout}, respectively, which are among the many sub-classes of \emph{trout}. Further, it should be noted that these hierarchies organize such classes as nouns in a taxonomy of increasing conceptual specificity, with \emph{Thing} or \emph{Entity} as their root \cite{miller1990}.

A few observations. First, the definitions of substance concept and of classification concept are very different, where the former is provided in terms of temporal sequences of frames, e.g., 2D or 3D videos, while the latter is provided in terms of linguistic descritpions, e.g., in terms of \textit{glosses} articulating Genus and Differentia \cite{miller1990}. Second, 
the coherence between what is visually represented by substance concepts and linguistically described by classification concepts is guaranteed by the fact that both types of concepts are ultimately generated from the same substances. Third, and most importantly, notice how there is a \emph{many-to-many mapping} between substances and substance concepts, between substances and classification concepts, and finally, by entailment, between substance concepts and classification concepts. To take an example, the substance \emph{chinook salmon} can be recognized as such or as a ``salmon", a ``fish", and vice-versa, depending upon the context, focus or purpose. Similarly, the substance fish can be reasoned about as a pet or as a food fish according to its contextual denotation in a lexical hierarchy.  Applying the approach introduced in \cite{giunchiglia2021a}, in this work these many-to-many mappings will be mapped to one-to-one mappings, as represented in the resulting concept hierarchies, by enforcing a correspondence, possibly but not necessarily with the user supervision, between the visual properties of substance concepts and the linguistically defined properties of classification concepts.

\section{A Knowledge Organization View of Concepts}
\label{sec:A Knowledge Organization View of concepts}
The teleological view of concepts gives us a functional categorization of concepts. The next issue is how to define a methodology which allows us to build hierarchies which are coherent with the teleological view. This is where Ranganathan's work takes a crucial role \cite{srr67}.
Following his approach, we organize concepts according
to the \emph{analytico-synthetic} paradigm. Such a paradigm, for any domain of discourse, derives its power from its two component procedures- \emph{analysis} wherein ideas are broadly recognized and decomposed into elemental facets which subsequently undergo \emph{synthesis}, involving the semantic composition of appropriate facets to form concepts. See \cite{giunchiglia2014} for a KR formalization of the analytico-synthetic paradigm. Within this approach, of specific interest to this work is the stratified mechanism proposed by Ranganathan \cite{srr67} which conjoins the perceptual recognition of concepts with their lexical-semantic expression and organization, grounded in the analytico-synthetic paradigm. The main novelty relates to the fact that this \emph{``separation facilitates the understanding and exploitation of each plane"} (quote from \cite{srr67}). We have the following phases (that Ranganathan calls also \textit{planes} or \textit{stages)}:

\begin{enumerate}
    \item \emph{Pre-Idea Stage}, which is focused on the perceptual generation of concepts;
    \item \emph{Idea Plane}, which is focused on the organization of concepts in a classification hierarchy, based on their perceptual (e.g., visual) properties;
    \item \emph{Verbal Plane}, which is focused on the  lexical-semantic rendering of the classification hierarchy (i.e., on linguistically naming the concepts); and,
    \item \emph{Notational Plane}, which is focused on formally rendering the classification hierarchy in language-agnostic terms, employing a unique numerical identifier for each concept in the hierarchy.
\end{enumerate}
The \emph{Pre-Idea Stage} is the phase which focuses on the \emph{cognitive grounding} of concepts via the process of their perception, recognition and subsequent mental agglomeration. Accordingly, we take perception to be the \emph{``reference of a percept to its entity-correlate outside the mind"} \cite{srr67}, and define two kinds of percepts facilitating incremental recognition - \emph{Pure Percepts} and \emph{Compound Percepts}. Pure Percepts are, quoting from \cite{srr67}, \emph{``a meaningful impression produced by any entity through a single primary sense and deposited in the memory"}, and Compound Percepts are \emph{``the impression, deposited in the memory, as a result of the association of two or more pure percepts formed simultaneously or in quick succession"} \cite{srr67}. To illustrate using an example, the (machine) visual acuity of a visual recognition system recognizes the impression produced by a fish eating a shrimp, where the impression produced of eating a shrimp is the \emph{pure percept}, and the object fish corresponds to the \emph{entity-correlate}. The same system, in a successive set of encounters, recognizes the impression produced by the same fish, but this time eating flake food (thus forming a different \emph{pure percept}). The system associates the two impressions together to form the compound percept fish in its memory, which is what we refer to as the \emph{``formation, deposited in memory, as a result of the association of percepts - pure as well as compound - already deposited in memory"} \cite{srr67}. The process of incremental assimilation of such \emph{``newly received percepts and newly formed concepts with the concepts already present in the memory"}, is what we call, from \cite{srr67}, \emph{apperception}, and the agglomerated memory which is characteristically in continuous evolution across encounters is referred to as \emph{apperception mass}. As from \cite{ceor,giunchiglia2021a}, in our terminology, a pure percept is the set of visual objects perceived during an encounter, a compound percept is the result of multiple encounters with the same substance, and apperception mass is the cumulative \textit{memory} of what has been perceived so far.

The \emph{Idea Plane}, being \emph{``a paramount plane which is both a map and foundation"} \cite{satija2017}, is built over the apperception mass through \emph{perceptual organization} of the perceived concepts, which Ranganathan terms \emph{Ideas} \cite{srr67}. Such perceptual organization is pragmatically effectuated by constructing \emph{perceptual subsumption hierarchies},  where these hierarchies correspond to the \emph{visual subsumption hierarchies} of visual concepts defined in \cite{giunchiglia2021a}. These hierarchies are formed by organizing such concepts following Genus-Differentia paradigm computed in terms of properties as perceived from objects. The design of such hierarchies is not \emph{based on intuition} but informed by a \emph{``a panoply of canons and postulates for designing and evaluating classification systems"} \cite{satija2017}. To illustrate with an example, when a visual recognition system will encounter a successive stream of visual frames composed of different aquatic animal-objects, it will be able to organize them (i.e., visual concepts induced by images) into a visual subsumption hierarchy  by forming genus and differentia in terms of their visual properties, and most importantly, guided by a set of established principles for rendering them ontologically thorough. As from \cite{ceor,giunchiglia2021a}, in our terminology, the Idea Plane corresponds to the organization of substance concepts into the visual subsumption hierarchy.

The \emph{Verbal Plane} employs \emph{``an articulate language as medium for communication"} \cite{srr67} of the concepts which are still in the form of a `perceptually-grounded' concept hierarchy. The crux of this phase is to seamlessly annotate such concepts (for instance, visual concepts in the form of objects in images) by employing semantically equivalent linguistic labels (mostly, nouns) from any number of natural languages or domain-specific vocabularies,\footnote{That is, an \emph{Object Language} as it is called in \cite{srr67}.} including also namespaces, and thus, in effect, assigning language label(s) to each such concept. As from \cite{ceor,giunchiglia2021a}, in our terminology, the Verbal Plane is the visual subsumption hierarchy transformed into a lexico-semantic classification concept hierarchy by labeling all the substance concepts with linguistic labels, articulating their Genus and Differentia, with respect to the other substance concepts. Notice that, as from \cite{fausto2017}, there will be a different hierarchy for each distinct natural language (e.g., English or Italian or Hindi) and that these hierarchies do not necessarily have the same shape, because of multilingual \textit{Lexical Gaps}. 


A consequence of the process of the linguistic annotation of the Verbal Plane, linguistic phenomena such as \emph{homonyms} and \emph{synonyms} get created, a fact which \emph{``causes aberration in communication"} \cite{srr67} and should be mitigated.  
This motivates the fourth and final plane, the \emph{Notational Plane}, which prescribes that language labels should be, quoting from \cite{srr67}, \emph{``replaced by symbols pregnant with precise meaning"} thus formally encoding the \emph{``uniqueness of the idea ... and the total absence of homonyms and synonyms"}. As from \cite{fausto2017}, and articulated in detail in Section \ref{sec:A Multimedia UKC}, the \emph{Notational Plane} is the true space of alinguistic concepts, uniquely identified, and organized into a classification hierarchy. The key observation is that, in the case of a multilingual hierarchy, as it is the case in the work in \cite{fausto2017}, because of lexical gaps, this hierarchy is a superset of the hierarchy associated to each and any single natural language.

\begin{table}[]
\caption{Mapping between Knowledge Organization and Teleological View of Concepts}
\label{tab:my-table}
\begin{tabular}{|l|l|}
\hline
\textit{\textbf{Knowledge Organization View}} & \textit{\textbf{Teleological View}} \\
\hline
Pre-Idea Stage   & Substance Concept Generation \\
Idea Plane       & Substance Concept Hierarchy     \\
Verbal Plane     & Substance Concept Hierarchy in words   \\
Notational Plane & Substance and Classification Concept Hierarchy \\
\hline
\end{tabular}
\label{table:kot}
\end{table}

The correspondence between Ranganathan's four-phased logical view of concepts and the teleological view of concepts, as detailed above, is represented in Table \ref{table:kot}. The order from top to down explicitly indicates how, progressively, what is being perceived is transformed into a hierarchy of classification concepts.
%
The key overall observation is the central role of substance concepts, and therefore of perception and visual properties, in particular during the first two phases, i.e., the pre-idea stage and the idea plane, where all the decisions about the organization of concepts are taken. This is fully coherent with the work described in \cite{giunchiglia2021a} where the hierarchy is built using substance concepts and where classification concepts, more precisely, the wording which describes them, are used only to linguistically label substance concepts. Let us elaborate on this fact.
The \emph{first} observation is that  Ranganathan's approach imposes a hierarchical organization of substance concepts, which, logically facilitates their mapping to classification concepts. The \emph{second} observation is that the notational plane inherits the substance concept hierarchy as built in the idea plane, where, quoting from \cite{satija2017}, the idea plane \emph{``genetically determines the quality of the ultimate product"}, i.e., the notational plane. Ranganathan characterizes, quoting from \cite{srr67}, \emph{``the relation between the idea plane and the notational plane"} as being \emph{``the one between a master and a servant"}, which is aligned with our own characterization of idea plane as \emph{the determiner of the taxonomic backbone of the (final) classification concept hierarchy}. The \textit{third} crucial observation is that the distinction between substance concepts and classification concepts is logically realized by applying, quoting from \cite{srr67}, \emph{``the Wall-Picture Principle ...} where \textit{... `Idea first, word next'"}. The intuition is that, just like a mural cannot be executed in the absence of a wall, there is no existence of (linguistically rendered and subsequently numerically identified) classification concepts without recognition of substance concepts in the first place. \textit{Fourth}, the very stratification of the process of building concept hierarchies aids \emph{``to solve independently, in the first instance, the problems arising in each plane"} \cite{srr67}, thus rendering each phase \emph{characteristically autonomous} yet \emph{functionally linked}. \emph{Finally}, it is worth noticing that the four phased mapping above is conceptually governed by the \emph{Law of Local Variation}, which is the principle that \emph{``there should be provision ... for strictly local use, results alternative to those for general use"} \cite{srr67}. This principle is crucial as it accommodates the fact that the mapping between substances, substance concepts and classification concepts can vary depending on, e.g., the purpose or focus.

Last but not least, notice how the process highlighted in Table \ref{tab:my-table} enforces the one-to-one mapping between substance concepts and classification concepts mentioned in Section \ref{sec:A Teleological View of Concepts}:  the proper natural language label and description will be selected based on the current (partial) view of the object under consideration. So, for instance the \textit{same} substance will be named, e.g., a \textit{person}, a \textit{woman}, \textit{Mary}, depending on the visual details which are perceived. In other words, the many-to-many mappings existing between substances, substance concepts and classification concepts mentioned at the end of Section \ref{sec:A Teleological View of Concepts} is properly encoded as a set of one-to-one mappings built by assigning labels not in terms of substances as such but, rather, in terms of the relevant substance concepts. It is worth noticing that this approach provides  a solution to a long standing unsolved problem that computer vision systems have, the so called \textit{Semantic Gap problem}, which was already identified  in 2010 \cite{smeulders2000} as (quote) ``\textit{... the lack of coincidence between the information that one can extract from the visual data and the interpretation that the same data have for a user in a given situation.}". In this quote we take substance concepts to encode `the information that one can extract from the visual data' and classifications concepts to encode `the interpretation that the same data have for a user in a given situation'. Because of how they have been constructed, all hiearchies of media constructed so far, including ImageNet \cite{imagenet2009}, suffer from the Semantic Gap problem (see also Section \ref{sec:A Multimedia UKC}).

\section{A Canonical Framework for Concept Hierarchies}
\label{sec:A Canonical Framework for Concept Hierarchies}
The adoption of Ranganathan's methodology enables us to exploit its normative principles, called \textit{canons},  which norm how to dynamically perform knowledge classification \cite{srr67}. The stress is on, quoting from \cite{srr89}, a \emph{``well designed classificatory language ... capable of individualising microscopic thought-units"}, thus facilitating \emph{``the representation of a multi-dimensional continuum in a single dimension"}. The pre-idea stage is not governed by canons as it is pre-eminently a \emph{causal phase} in generating new concepts from objects via recognition. We analyse below other three phases where, as to be expected, the canons for the idea plane are by far the most important.

\subsection{Canons for Idea Plane}

The canons of the idea plane are organized in four specialized sets, to be applied sequentially one after the other. They are: (i)
\emph (canons about) {characteristics}; (ii) (canons about) \textit{succession of characteristics}; (iii) (canons about) \textit{arrays} and (iv) (canons about) \emph{chains}. Let us analyze them in detail.

\vspace{0.1cm}
\noindent
\emph{Characteristics} (by which we mean, outer characteristics, in our terminology, substance properties) form the basis of classification of substance concepts, and the objective is to select such characteristics as will be helpful for our purpose. Let us consider the four which are most relevant.
The \emph{canons of differentiation} and \emph{relevance} are conjoined in their purpose, in the sense that the former ensures that a characteristic employed for classifying substance concepts should, quoting from \cite{srr67}, \emph{``differentiate some of its entities- that is, it should give rise at least to two classes"}, whereas the later corroborates that such a characteristic \emph{``should be relevant to the purpose of the classification"} itself \cite{srr67}. For example, while the impossibility of unambiguous classification of salmon and trout on the basis of (visual) recognition of gills in them is ensured by the canon of differentiation, relevance informs that fin spots are an appropriate (visual) differentiating characteristic if the purpose is to classify fishes as per geographical habitat. 
Further, the \emph{canon of ascertainability} enforces that a classifying characteristic \emph{``should be definite and ascertainable"} \cite{srr67}, in perceptual terms. To take an example, the presence or absence of \emph{pyloric caeca}, which is a part of internal anatomy in many fishes, cannot be construed as an ascertainable characteristic for visually classifying different fishes. 
Finally, the \emph{canon of permanence} states that, quoting from \cite{srr67}, \emph{``a characteristic used as the basis for classification ... should continue to be unchanged, so long as there is no change in the purpose of classification"}, a direct exemplar of which is the fact that \emph{colour} cannot be used as a (perceptual) classificatory characteristic for those \emph{fishes which camouflage}.

\vspace{0.1cm}
\noindent
The next step is the \emph{succession of characteristics}, namely the order by which characteristics should be applied. It is important to notice that this ordering is crucial as, in case of shared properties, different orderings generate different hierarchies. As an example, the \emph{canon of relevant succession} posits, quoting from \cite{srr67}, \emph{``the succession of the characteristics ... should be relevant to the purpose of the classification"}. To illustrate, let us take the case of a visual recognition application for recognizing different fishes. The first logical (visual) characteristic to differentiate, for instance, between salmon and trout will be the \emph{tail shape}, with respect to which the former has a concave tail whereas the trout's tail is convex shaped. Further, the presence or absence of \emph{round parr marks} can be used by the application as the second (visual) characteristic to differentiate between different varieties of trout, such as rainbow trout and steelhead trout.

\vspace{0.1cm}
\noindent
The progressive application of the canons for characteristics and succession of characteristics leads to the formation of \textit{arrays}, which are groups of classes, or categories, bearing coordinate status (i.e., categories which are children of the same node), at all the levels of the subsumption hierarchy. Such formation of arrays are guided by the \emph{canon of exhaustiveness}. Exhaustiveness mandates that classes belonging to an array, quoting from \cite{srr67}, \emph{``should be totally exhaustive of their respective common immediate universes"}, and further, \emph{``any new entity added to the original universe ... should be assigned to any of the existing classes or to a newly formed class"}. This is crucial for visual recognition applications where, for example, all the known varieties of salmon should be made coordinate subclasses of the class salmon with the possibility that a newly discovered variety of salmon can be assigned to any of the existing classes or be classified as a new one based on the recognition of a new set of \emph{visual properties}. 

\vspace{0.1cm}
\noindent
The last step is the formation of \textit{chains}, namely what in graph theory are called \textit{paths}.
Here the two canons which are pivotal for developing \emph{taxonomically clean} chains are the canon of \emph{increasing/decreasing extension} and the canon of \emph{modulation}. The canon of increasing/decreasing extension is centered around the correlative notions of extension which \emph{``measure the number of entities or of the range comprised in the class"} \cite{srr67}, and intension which signifies the properties that can be predicated of it. Based on these notions, decreasing extension states that while traversing down a chain, quoting from \cite{srr67}, \emph{``from its first link to its last, the extension of the classes ... should decrease and the intension should increase at each step"}. Increasing extension, on the other hand, conveys the exact opposite in case of upward traversal in a chain.
The second and last canon that we consider is the canon of \textit{modulation} which states that such a chain should comprise one class \emph{``of each and every order that lies between the orders of the first link and the last link of the chain"}, or in other words, the assertion that a chain shouldn't have any \emph{missing link}. A direct consequence of this canon on the ability of recognition (especially for \emph{human-like visual recognition}) is exemplified by the established fact \cite{rosch1976} that there are certain basic categories that are probabilistically most optimal to be perceptually recognized and can never be missed out (for example, for fishes, we cannot skip the class fish and directly jump from aquatic vertebrate to salmon). 

Notice that  while some of the canons mentioned above are more or less always followed in the state of the art (linguistically constructed) hierarchies, others are not, thus resulting in classification of low quality. Some examples in the first class are the canon of differentiation and the canon for increasing/ decreasing extension which holds by construction in all hierarchies  built using Genus-Differentia. Examples of the second class are: the canon of permanence, the canon of relevant succession, (sometimes) the canon of exhaustiveness and the canon of modulation.

\subsection{Canons for Verbal Plane}

The next step is to linguistically label substance concepts with language labels (nouns). The \emph{canon of context} prescribes, from \cite{srr67}, \emph{``the denotation of a term in a scheme for classification should be determined in light of the different classes (Upper links) ... belonging to the same primary chain as the class"}. It is unified in its purpose with the \emph{canon of enumeration} which stipulates such a denotation to be also determined \emph{``through the subclasses ... enumerated in the various chains having the class ... denoted by the term in question as their common link"} \cite{srr67}. The two canons above logically mediates the many-to-many mapping between substance concepts and classification concepts. To take an example, the contextual recognition of the substance concept `fish' as an aquatic vertebrate, a pet or a food depends on its superordinate classes, whereas its precise extensional meaning, in other words its sense disambiguation in classification concept terms, is defined by the subclasses it enumerates in the context of the linguistic hierarchy. Finally, the \emph{canon of reticence} states, from \cite{srr67}, \emph{``the term used to denote a class ... should be the one current among specializing in the subject field"}, or in other words, it prescribes the usage of an appropriate domain language (such as namespaces) for unambiguous annotation of substance concepts, for instance, images. The main goal here is to avoid the use of synonyms.

Here it is to be noted that all these canons are most often followed in the state of the art hierarchies, the first holding by construction in Genus-Differentia hierarchies, the second holding any time the \cite{get-specific} principle is applied.

\subsection{Canons for Notational Plane}

The canons for the notational plane are aimed at translating the linguistic hierarchy of the verbal plane into a fully formal hierarchy of alinguistic classification concepts, wherein each concept (more specifically, each sense) is associated to a unique numerical identifier. The \emph{canon of synonym} specifies that, quoting from \cite{srr67}, \emph{``each isolate idea should be represented by one and only one isolate number"}, which, in our context, ensures that each classification concept is representated by one and only one identifier. On the other hand, the \emph{canon of homonym} implies that  \emph{``each isolate number should represent one and only one isolate idea"} \cite{srr67}. Thus, these two canons, in effect, impose a \emph{necessary and sufficient} condition between concepts and their respective identifiers. Further, the \emph{canon of hospitality in arrays and chains} \cite{srr67}, for us, states that a new concept can be appropriately positioned and uniquely identified anywhere in the hierarchy. These canons cumulatively ascribes to the notational plane the quality of \emph{perpetuation}, \emph{``the devices necessary and sufficient to represent uniquely and unequivocally—that is, to individualize—every new formation thrown forth ... from time to time"} \cite{srr49}, which attests to its continuous evolution. Thus, a true \emph{classification concept hierarchy} emerges in the Notational Plane, with the unique identifiers performing Word Sense Disambiguation (WSD) and also rendering the space, synonym, homonym and polysemy free, at the same time. 

These canons, while not holding in general, are satisfied by all WordNet-like hierarchies \cite{miller1990}.

\section{From media to classification concepts via substance concepts}
\label{sec:A Multimedia UKC}
The use of media, e.g., videos or photos, is quite pervasive, in particular, but not only, in the Web. This phenomenon extends also to hierarchies, for instance, in the case of eCommerce, where the user is able to seamlessly navigate a catalog where each item is annotated by, usually, an image. The key observation is that, in these situations, the main description is provided in natural language, while photos have a complementary role of integrating visually the main content provided linguistically. ImageNet \cite{imagenet2009} is a very important point in case. ImageNet is a very large image database which is extensively used for the training of Deep Neural Networks. It has been built by taking WordNet, its English Version from Princeton \cite{miller1990}, and by populating it with millions of photos collected from the Web. As also described in \cite{imagenet2009}, the construction of ImageNet has been done in a way to preserve a high level of quality. However, for how it has been constructed, \textit{viz.} by populating a linguistic hierarchy with photos, there is no guarantee that the photos provide the information that would be needed to build the visual subsumption hierarchy implicitly assumed by WordNet. In other words, while by construction, ImageNet is a hierarchy of classification concepts, there is no evidence that the photos encode also substance concepts. While this will most likely have no implications in the recognition of single substances, some difficulties may arise in case one is interested in learning concept hierarchies, and also any time the problem of the Semantic Gap raises difficulties (see discussion in Section \ref{sec:A Knowledge Organization View of concepts}). Furthermore notice how these limitations apply to all approaches where linguistic hierarchies are used to classify objects, see, e.g., 
\cite{shvor,labelme} or, more generically, to support computer vision, see, e.g., \cite{ontocv}.

The goal of the project introduced in this section, whose preliminary name is \textit{MultiMedia UKC}, is to build a resource very similar to ImageNet but with the key difference of being built following the methodology defined in this paper. The starting point is the Universal Knowledge Core (UKC) \cite{fausto2017},\footnote{An online version of the system can be reached at \url{http://ukc.datascientia.eu/}.} a multilingual lexical resource now containing more than one thousand languages and more than one hundred thousand classification concepts. Besides its size, both in terms of concepts and languages, which is a strong incentive towards its use, the UKC seems very well suited for our goals as its organization matches quite naturally Ranganathan's four-phased methodology. Starting from the UKC, we envision the following construction of the MultiMedia UKC:

\begin{itemize}

\item \textit{Pre-Idea Stage}: This phase is used to construct (substance) concepts by extracting \textit{visual objects} from media. The media will be selected to depict the terms in the UKC. The extraction of visual objects will be done applying the techniques from \cite{ceor,giunchiglia2021a};

\item \textit{Idea Plane}: This phase is used to \textit{construct} a visual subsumption hierarchy via visual objects. This is done by applying the methodology described in this paper;

\item \textit{Verbal Plane}: This phase is used to annotate substance concepts with \textit{words}, which, in turn, are annotated with \textit{synsets} \cite{miller1990}, i.e., the set of their synonyms. Synsets are further annotated with their definition (i.e., their \textit{gloss}) defined  in terms of Genus and Differentia. There is one verbal plane per language. This is achieved by aligning the hierarchy constructed in the idea plane with the UKC hierarchy;

 \item \textit{Notational plane}: This phase is used to generate a set of language independent classification concepts, as unique, alinguistic identifiers. This is done reusing the concepts which already exist in the conceptual layer of the UKC.
 
\end{itemize}
\noindent
The construction of a multimedia UKC is quite ambitious and it is bound to raise complications in all the first three steps. The notational plane should come for free by reusing the current UKC identifiers. In the pre-idea stage the complexity comes from the issue of how to extract features from media, a problem which can get rather complicated in non-ideal situations, e.g., in presence of noise. The intuition is to solve this problem with the help of the human supervision. In the Idea plane the complication comes from the need of selecting how to apply the canons introduced in Section \ref{sec:A Canonical Framework for Concept Hierarchies}, a task which requires the involvement of an expert. The complication in the verbal plane comes from the need to match the hierarchy built in in the previous step with the hierarchy which already exists in the UKC. 
In practice we expect the classification process of the second step to be strongly driven by what is already available in the UKC. And for sure there will be a lot of iterations between these two steps. 

To illustrate the kind of reasoning that we will have to perform while constructing the MultiMedia UKC, we introduce two small examples, which relate to two key tenets of categorization from the work on basic categories by  Eleanor Rosch \cite{rosch1976,rosch1999}. The starting point is her empirical observation that in taxonomies (quote from \cite{rosch1976}),

\vspace{0.1cm}\noindent
\emph{``there is one level of abstraction at which the most basic category cuts are made"}, where she defines basic categories as \emph{``those which carry the most information ... and are, thus, the most differentiated from one another"}.

\vspace{0.1cm}

\vspace{0.1cm}\noindent
Coherently with the methodology proposed here, she observes that in perceiving and categorizing objects (quote from \cite{rosch1999}), 

\vspace{0.1cm}\noindent
\emph{``objects may be first seen or recognized as members of their basic category, and that only with the aid of additional processing can they be identified as members of their superordinate or subordinate category."} 

\vspace{0.1cm}
\noindent
In complement to the above tenet, her second tenet  establishes that (quote from \cite{rosch1999})

\vspace{0.1cm}\noindent
\emph{``basic objects appeared to be the most abstract categories for which an image could be reasonably representative of the class as a whole"}.

\begin{figure}[htp]
    \centering
    \includegraphics[width=12cm, height=5cm]{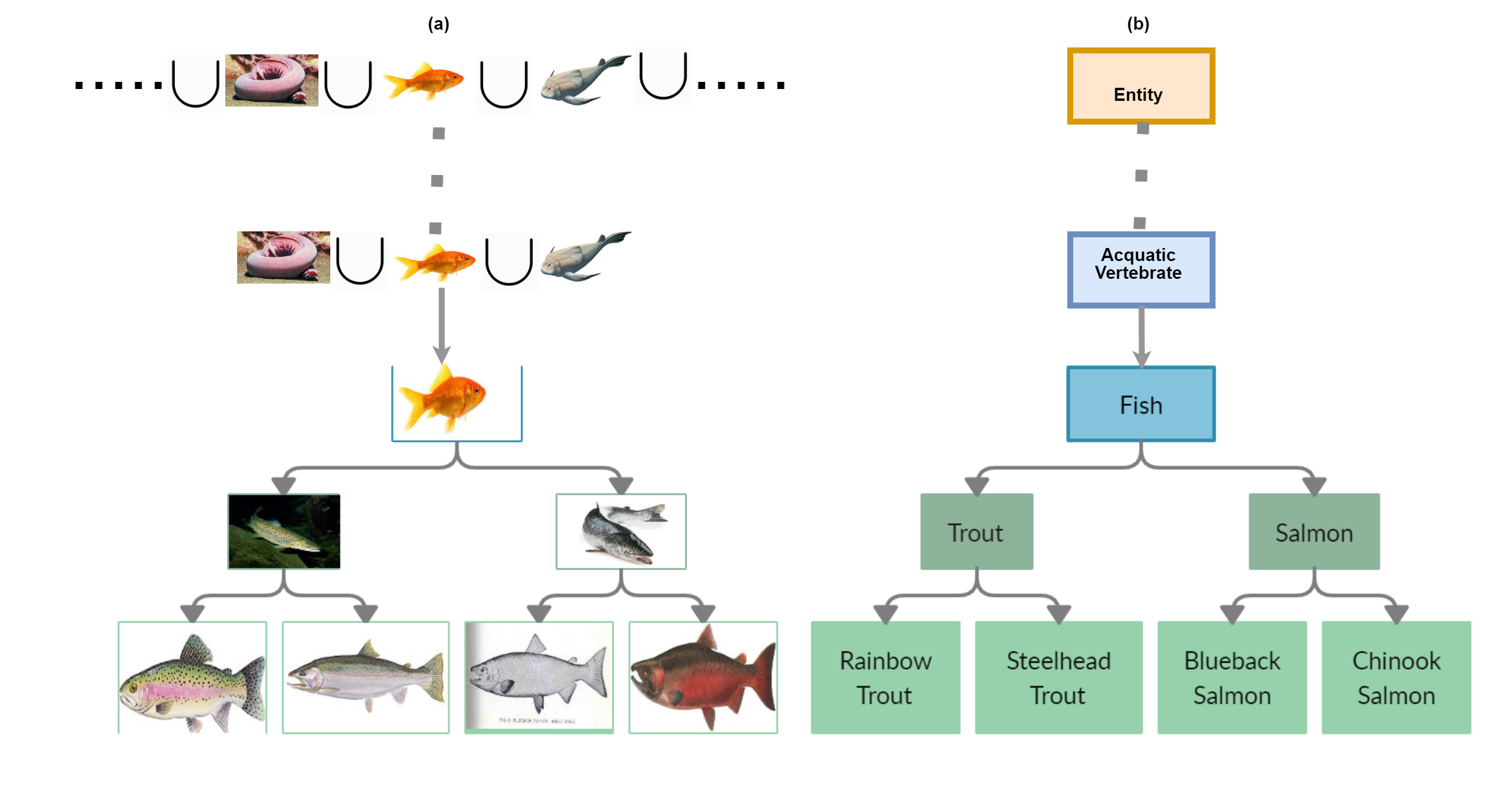}
    \caption{\textbf{Biological Taxonomies of Fish (picturized from Rosch \textit{et al.} \cite{rosch1976})}}
    \label{fig:FOIS2021BTU}
\end{figure}
\noindent
Let us concentrate on the biological taxonomy depicted in Figure \ref{fig:FOIS2021BTU}, as from \cite{rosch1976}, where the second classification is a natural language description of the first. Following Rosch's first tenet, \emph{fish} is a basic category. In fact, fishes share the maximum number of visual properties amongst themselves and are also most differentiable amongst other sub-categories of \emph{aquatic vertebrate} (e.g., \emph{placoderm} and \emph{agnathan}, not mentioned in the linguistic hierarchy on the right of Figure \ref{fig:FOIS2021BTU}, but depicted in the left figure by the images which annotate the nodes above the node associated to fish). 

The first example considers the situation when we move upwards from the basic category \textit{fish}, therefore increasing extension. Here the role of the canon of increasing/decreasing extension provides the logical means of organizing the superordinate categories into a taxonomically clean chain. 
Let us consider for instance the concept of  \emph{aquatic vertebrate}. In Figure \ref{fig:FOIS2021BTU} the big \textit{union symbol} sign means that the extension of this concept should be taken as the disjoint union of the elements depicted.
As the picture suggests, this concept  cannot be visually recognized by any representative image and can only be perceived by considering the images of all the different basic categories, e.g., \emph{fish}, \emph{placoderm} and \emph{agnathan}. In other words more abstract substance concepts should be constructed by joining the substance concepts of the child concepts. But this comes for free, and the distinction is only linguistic and \textit{not visual} as, also for the lower categories, e.g., fish, substance concepts are the union of distinct and different visual objects. 
This is a general phenomenon which we believe it will apply any time we will move towards the root of the hierarchy and that, it seems, has been largely overlooked so far in the mainstream computer vision literature. Another example is the concept of \textit{vehicle}, which can only be visualized as the union of the extension of its subordinate concepts, some of which are basic categories, e.g., \textit{car}, \textit{bike}, \textit{ship}, \textit{train}.
Dually, the subordinate categories of \textit{fish}, e.g., \emph{blueback salmon} \emph{chinook salmon}, on the other hand, can be visually recognized by incrementally recognizing the finer visual properties of the basic category \emph{fish} (decreasing extension) over successive encounters. However how far down it is possible to go in the recognition of the subordinate categories before falling into Rosch's second tenet, is an open question for which, at the moment we have no answer. We expect that it will vary a lot from one basic category to another.

The canon of modulation, on the other hand, logically facilitates the identification of co-extensiveness of the categories which are superordinate and subordinate to the basic categories by ensuring the impossibility of missing links. As a matter of fact, this factor definitely confirms the primacy of  basic categories in the process of perception. For example, a  recognition of \emph{rainbow trout} as a subordinate category of \emph{fish} thus skipping \emph{trout} as the category between \textit{fish} and \textit{raimbow trout} fails in its very purpose of incremental visual classification as it will bring up \emph{rainbow trout} to the same level of visual co-extensiveness as \emph{salmon}. Further, if the system visually recognizes \emph{trout} to be the basic category and \emph{aquatic vertebrate} to be its immediate superordinate category, the system fails in its adherence to human-like vision which, as has been extensively established in \cite{rosch1976,rosch1999}, recognizes \emph{fish} as a biological basic category. Thus, to restate, the impossibility of missing links (and hence, the canon of modulation) ensures the primacy of basic categories. This at the moment is only an intuition. We believe that the construction of the MultiMedia UKC will allow to (dis)confirm this intuition quantitatively.


\section{Conclusion}
\label{sec:Conclusion}
Usually, when implementing object recognition systems, and also when building lexico-semantic hierarchies, media, e.g., videos or photos, are organized and classified based on their linguistic description. This is correct as the purpose of language is exactly that of describing what is known.
However, based on some recent results in Computer Vision, this paper suggests that, when in the process of recognizing objects and classifying them in conceptual hierarchies, visual properties are much more relevant. It also suggests to use Ranganathan's faceted approach which is articulated exactly in terms of how visual properties should be progressively refined  up to the generation of a not ambiguous linguistic description. The work described here are just the first steps towards our ultimate goal, namely the construction of a large scale multilingual multimedia lexical resource.

\section*{Acknowledgements}

The research conducted by Fausto Giunchiglia, and Mayukh Bagchi has received funding from the \emph{“DELPhi - DiscovEring Life Patterns”} project funded by the MIUR Progetti di Ricerca di Rilevante Interesse Nazionale (PRIN) 2017 – DD n. 1062 del 31.05.2019.

\end{document}